\documentclass{article}
\usepackage{arxiv}
\usepackage[utf8]{inputenc} 
\usepackage[T1]{fontenc}    
\usepackage{hyperref}       
\usepackage{url}            
\usepackage{booktabs}       
\usepackage{amsfonts}       
\usepackage{nicefrac}       
\usepackage{microtype}      
\usepackage{lipsum}		
\usepackage{graphicx}
\usepackage{natbib}
\usepackage{doi}
\usepackage{float}

\title{Towards an approach to multivariate outlier detection for District Heating System data}

\date{} 					

\author{ \href{https://orcid.org/0000-0002-6324-5967}{\includegraphics[scale=0.06]{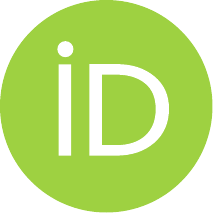}\hspace{1mm}Rajko Turudija}\thanks{This research was supported by the Science Fund of the Republic of Serbia, Grant No. 23-SSF-PRISMA-206, Explainable AI-assisted operations in district heating systems - XAI4HEAT} \\
	Faculty of Mechanical Engineering\\
	University of Niš\\
	ul. Aleksandra Medvedeva 14, 18000 Niš, Serbia \\
	\texttt{rajko.turudija@masfak.ni.ac.rs} \\
	\And
	\href{https://orcid.org/0000-0002-6787-4533}{\includegraphics[scale=0.06]{orcid.pdf}\hspace{1mm}Dušan Stojiljković} \\
	Faculty of Mechanical Engineering\\
	University of Niš\\
	ul. Aleksandra Medvedeva 14, 18000 Niš, Serbia \\
	\texttt{dusan.stojiljkovic@masfak.ni.ac.rs} \\
    \And
	\href{https://orcid.org/0000-0002-4141-2058}{\includegraphics[scale=0.06]{orcid.pdf}\hspace{1mm}Milan Zdravković} \\
	Faculty of Mechanical Engineering\\
	University of Niš\\
	ul. Aleksandra Medvedeva 14, 18000 Niš, Serbia \\
	\texttt{milan.zdravkovic@masfak.ni.ac.rs} \\
    \And
	\href{https://orcid.org/0000-0001-7205-7987}{\includegraphics[scale=0.06]{orcid.pdf}\hspace{1mm}Marko Ignjatović} \\
	Faculty of Mechanical Engineering\\
	University of Niš\\
	ul. Aleksandra Medvedeva 14, 18000 Niš, Serbia \\
	\texttt{marko.ignjatovic@masfak.ni.ac.rs} \\
}

\renewcommand{\headeright}{}
\renewcommand{\undertitle}{}

\hypersetup{
pdftitle={A template for the arxiv style},
pdfsubject={q-bio.NC, q-bio.QM},
pdfauthor={David S.~Hippocampus, Elias D.~Striatum},
pdfkeywords={First keyword, Second keyword, More},
}

\begin{document}
\maketitle
\begin{abstract}
	In this paper, we test different methods for multivariate detection of outliers in the data of transmitted heat energy in the selected substation of local District Heating System, by also considering outside ambient temperature, namely Z-score (univariate, as a benchmark), Mahalanobis distances, Principal Component Analysis (PCA), Isolation Forest and Hotelling's T-squared test. The overall research aims at uncovering irregular plant operation, with a wider objective of identifying the opportunities for reducing the consumption of gas in central heating plants as well as the CO2 emission. The proposed approach considers specific domain circumstances, such as irrelevance of zero transmit-ted energy timepoints as indication of off-grid plant. The outcomes of the different methods are discussed with domain experts. It was concluded that PCA, Isolation Forest and Hotelling method provide relevant results. Finally, we adopt the ensemble method (selection based on the agreement of all three methods on the detected outliers) as the final approach.
\end{abstract}

\bigskip
\noindent\textbf{Publication note.}
This preprint corresponds to the paper published as:

R.~Turudija, D.~Stojiljković, M.~Zdravković, and M.~Ignjatović,
``Towards an Approach to Multivariate Outlier Detection for District Heating System Data,''
in \emph{Disruptive Information Technologies for a Smart Society},
M.~Trajanović, N.~Filipović, and M.~Zdravković (Eds.),
Lecture Notes in Networks and Systems, vol.~860,
Springer, Cham, 2024.
The final authenticated version is available at
\url{https://doi.org/10.1007/978-3-031-71419-1_5}.

\section{Introduction}
District Heating System (DHS) plant operation refers to the automated or semi-automated management of water temperature and water flow in both the main plant and substation supply lines. This control is based on the overall demand for DHS and the prevailing weather conditions. The total DHS demand is considered by analyzing the transmitted energy per time period, as measured by the calorimeter at the return lines. Normally, this transmitted energy is correlated with the outside ambient temperature, which is used as a main control parameter in DHS plant operation.

DHS are designed to efficiently distribute heat to multiple buildings within a geographical area. Anomalies in data, such as sudden spikes or drops in energy consumption, can indicate inefficiencies or malfunctions in the system. Detecting these anomalies allows for timely intervention to optimize energy usage and minimize waste. Anomalies in data can highlight fluctuations in heating performance, which may result in customer complaints. Identifying and rectifying anomalies promptly enhances service quality and customer contentment. When looking for anomalies beyond the transmitted energy, data can reveal abnormalities in the performance of different components of DHS, such as boilers, pumps and pipes, which may be indicative of impending failures or maintenance needs. Anomaly detection techniques can be used to implement predictive maintenance strategies in district heating systems. By identifying anomalies in equipment or performance data, maintenance schedules can be optimized to reduce downtime and minimize costs. Detecting anomalies in sensor data or measurement errors helps maintain the accuracy and reliability of data used for decision-making and modeling.

Anomaly detection is the problem of highlighting unexpected items or events (differing from a norm) in datasets or data streams. It is often used on time-series data, especially in industries, for fault detection.  There are many different approaches for anomaly detection, from conventional (such as statistical ones, Isolation Forest, Principal Component Analysis, etc.) to advanced, based on DL architectures (instance-based anomaly detection \cite{Teng2010}, generative adversarial networks \cite{Li2019}, LSTM-based Encoder-Decoder architectures \cite{Malhotra2016} and others). Although the latter are much more effective, black-box nature of DL networks is considered as significant drawback for application in the industrial system, where the explanation of detected anomaly is crucial for root-cause analyzes and justification of high-impact shop-floor decisions (such as revision of operation parameters, equipment maintenance, etc.).

In this paper, we define the initial approach of general research to detect anomalies in data related to heat energy consumption in the local DHS. Our approach explores the feasibility of use of the traditional methods to univariate and multivariate outlier detection, namely z-score (univariate, as a benchmark), Mahalanobis distances, Principal Component Analysis (PCA), Isolation Forest, and Hotelling's T-squared test.
In section 2, we detail the problem and introduce some of the reference works in this area. In section 3, approach methodology and dataset are described, as well as candidate methods for outlier detection. In section 4, the implementation process is briefly described, after which we present a detailed discussion of the results with the adopted approach formulation. Finally, in section 5, conclusions and further work are presented.

\section{Background}
\label{sec:headings}

Traditional DHSs are currently governed by a Supervisory Control And Data Acquisition (SCADA) system. This system incorporates various sensors, control mechanisms, and integrated algorithms that adjust operational parameters based on sensor readings. The control of DHS is entirely automated, encompassing both the boiler plant and the district heating substation levels. It also involves the implementation of suitable hot water reset control strategies, such as outdoor air reset or control curves, often represented by a regulation curve with multiple control set points.

\begin{figure}[H]
	\centering
	\includegraphics[width=0.7\textwidth]{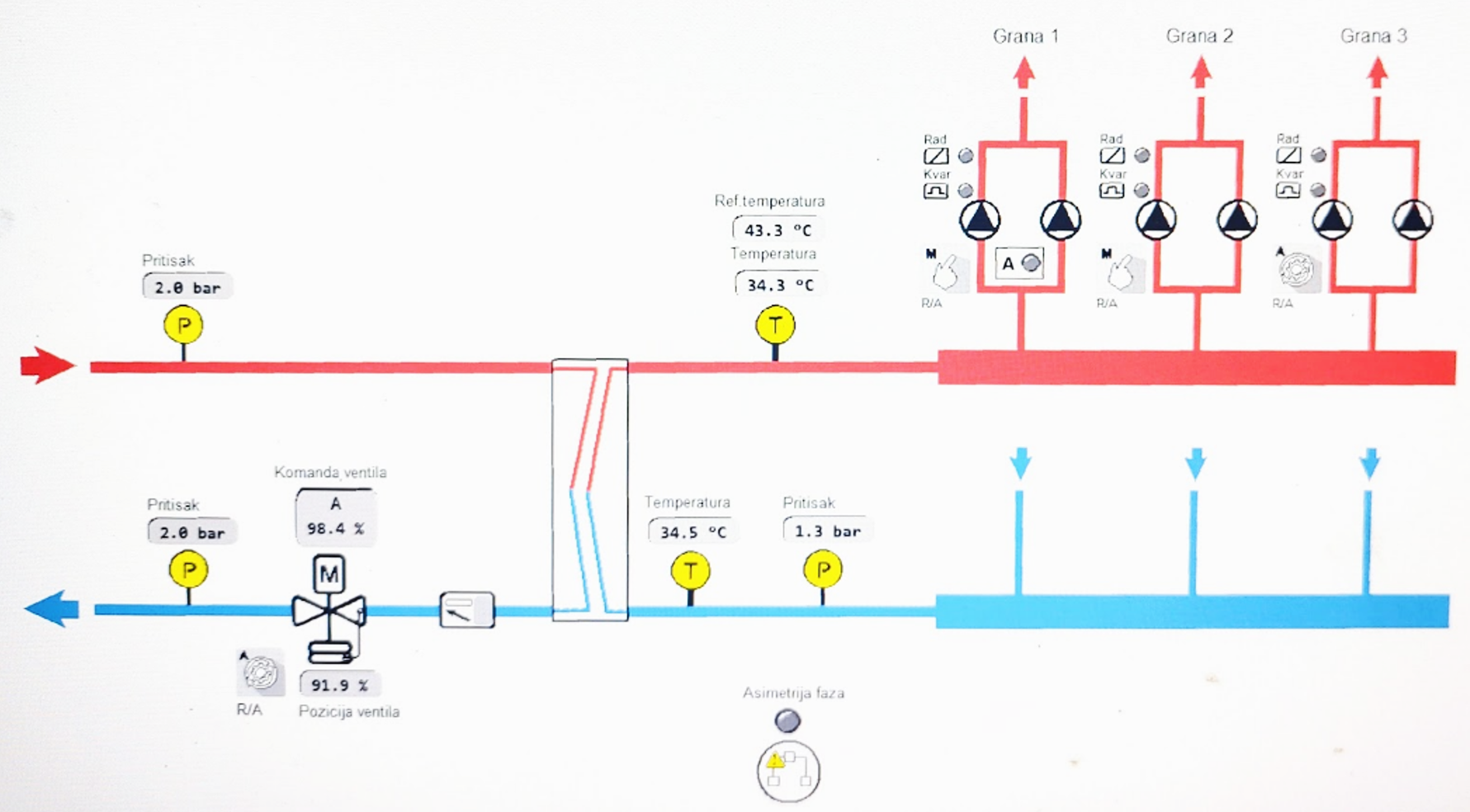}
	\caption{Substation 8 scheme}
	\label{fig:fig1}
\end{figure}

Selected substation scheme is presented in Fig.1. Heat fluid flow takes place in primary and secondary lines, separated by the heat exchanger. The central heating plant generates hot water or steam as a heat carrier. This high-temperature fluid carries thermal energy from the plant to the various consumers, namely the substations. This is a primary supply line. At the substation, the high-temperature fluid from the central plant passes through a heat exchanger. This heat exchanger is typically designed to transfer heat from the central fluid to a separate, closed-loop system, from primary to secondary supply line. In the closed-loop system, the fluid is pumped to the individual customer's heating system (through branches). This closed loop prevents the central heating fluid from mixing with the customer's internal system. After transferring its heat to the customer's closed-loop system, the temperature of the fluid in the substation decreases. This cooled fluid returns to the substation. This is a secondary return line. At the substation, the cooled fluid is reheated using heat from the central heating plant. The heat exchanger in the substation facilitates this process. Transmitted heat energy is measured at the calorimeter which is located at the primary return line. This energy is proportional to the difference between temperatures in secondary supply and return lines and flow in a secondary line.

\subsection{Detecting faults and anomalies in DHS}

Anomaly detection involves identifying patterns in data that deviate from the expected behavior, often referred to as outliers or aberrations \cite{Chandola2009}.

In data analysis, outlying observations can significantly impact results, emphasizing the need for robust statistical methods (or other methods) to identify and mitigate their influence. Robust statistics focus on detecting outliers by identifying the model that best fits most of the data. Rousseeuw and Hubert \cite{Rousseeuw2011} present an overview of various robust methods and outlier detection tools, addressing univariate, low-dimensional, and high-dimensional data. The discussion encompasses robust procedures for estimating location and scatter, handling linear regression, Principal Component Analysis (PCA), and classification, providing a comprehensive exploration of techniques to enhance the reliability of data analysis in the presence of outliers. On the other hand, a fault is characterized as a change in a system that renders it incapable of satisfactory operation and meeting user requirements \cite{Worden2004}. The distinction between detecting faults and anomalies in systems is highlighted by Neumayer et al \cite{Neumayer2023}.

Al Koussa and Månsson \cite{AlKoussa2022} present two methods for detecting faults in district heating (DH) substations using energy meter data. The first method is a cluster-based approach, comparing substations within a network to identify suboptimal performance. The second is an instance-based approach, using a black-box model to predict substation behavior and identify deviations from expected performance. Both methods demonstrate effectiveness in detecting deviating behaviors in DH customer installations, offering a significant advancement in automated fault detection in this field. This distinction underscores the need for precision in discerning irregularities in system behavior, whether they manifest as anomalies or faults.
Mbiydzenyuy and Sundell \cite{Mbiydzenyuy2022} address the challenge of unsupervised anomaly detection in District Heating, citing limited labeled data and complex structures. Their solution involves algorithm which employs hierarchical clustering, Dynamic Time Warping, Matrix profiles, and a Generative Adversarial Network to detect anomaly instances with similar patterns to labeled anomalies. The findings demonstrate a strategic approach for energy analysts to assess specific anomalies in the District Heating network, unveiling prevalent cluster patterns within the dataset.

Zhang and Fleyeh \cite{Zhang2020} introduce a novel hybrid approach for anomaly detection in DHS substations, acknowledging the common occurrence of system under-performance due to various faults. The proposed method combines a simplified physical model with a Long Short-Term Memory-based Variational Autoencoder (LSTM VAE). Using a dataset from an anonymous Swedish substation, the study evaluates and compares the performance of two state-of-the-art models, namely LSTM and Long Short-Term Memory-based Autoencoder (LSTM AE), with LSTM VAE. Results indicate that LSTM VAE outperforms baseline models when applying an optimal threshold.

Seem \cite{Seem2007} outlines an innovative approach for identifying abnormal energy consumption in buildings through daily energy readings and peak consumption data. Utilizing outlier detection, the method assesses whether daily energy consumption significantly deviates from previous levels. For buildings exhibiting abnormal energy patterns, robust estimates of mean and standard deviation quantify the variation from the norm. This data analysis technique promises cost reduction by detecting unnoticed issues and saving operational time previously spent on manual fault detection. Successful detection of high-energy consumption in various buildings is demonstrated through field test results, showcasing its efficacy in identifying issues such as chiller failure, suboptimal control strategies, ventilation system design flaws, and improper equipment operation following electrical panel changes.

Within the broader research scope aimed at revealing irregularities in plant operation and ultimately identifying opportunities to reduce gas consumption and CO2 emissions in central heating plants, this paper marks the initial step of the overall research in that trajectory. The paper delves into the exploration of various outlier detection methods, assessing their effectiveness. In collaboration with domain experts, the identified outliers are clarified through their expertise, ultimately determining the suitable approaches for accurate outlier detection in the pursuit of optimizing plant efficiency and environmental sustainability.

\section{Methodology}
\label{sec:headings}

Obviously, one daily operation is expected to follow the specific pattern. This pat-tern includes the peaks that are the result of the operator decisions and valleys which occur when the heating system is off. To avoid detection of above-mentioned peaks as anomalies, in our approach we detect anomalies on the number of datasets, for which data included only transmitted energy and ambient temperature at the given hour. For example, we will separately analyze all data collected at 7:00 and at other hours.
The approach is used on the data corresponding to the specific hours due to a very clear seasonality daily pattern for the transmitted heat energy corresponding to the human operator procedures. The outlier detection approaches will be tested on the daily time series comprising of the ambient temperature and transmitted heat energy at the following hours of a day: 7, 8, 9, 10, 11, 12, 13, 14, 22, 23. Detection will consider both time series, so the bi-variate outlier detection approach will be used. Multivariate outliers are data points that are unusual or extreme when considering multiple variables simultaneously. In other words, these are observations that are outliers in a multivariate space. As a reference, a simple statistical Z-score method will be used on transmitted heat energy signal.

\subsection{Dataset description}

The dataset includes following relevant features:
\begin{itemize}
    \item Outside air temperature, acquired by the sensor in the substation (C, tsp)
    \item Reference temperature (C, trt). Reference temperature is a target temperature at secondary supply line which needs to be reached. Reference temperature is calculated by the control curve.
    \item Water temperature in the secondary supply line (C, tns)
    \item Water temperature in the secondary return line (C, tps)
    \item Water temperature in the primary supply line (C, tnp)
    \item Water temperature in the primary return line (C, tpp)
    \item Heat energy transmitted (MWh, e)
\end{itemize}
Data is collected from substation 9 of the local DHS by merging different datasets. The data period is 2018-05-05 - 2023-05-01. Normally, the adopted time of the heating season start is the time of first change in transmitted heat energy after October 1st. The end of the heating season is normally April 15th. Although the heating system often remains operational after that date, till May 3rd, this data is discarded because of high temperature peaks that often occur in this period. Heat energy is measured at the calorimeter which is located at the primary supply line. Due to special conditions for DHS operation in October, this month will also be omitted from the analysis. For data analysis, the adopted period is 11-01 - 04-01.

\subsection{3.2	Multivariate outlier detection methods}

Z-score as the simplest but often most often univariate outlier detection technique will be used as a reference. Z-score is a statistical measure that quantifies how far is a data point from the mean of the dataset in terms of standard deviations.

\begin{equation}
    Z = \frac{X - \mu}{\sigma}
\end{equation}

where:
\begin{itemize}
    \item $X$ - individual data point.
    \item $\mu$ - mean (average) of the dataset.
    \item $\sigma$ - standard deviation of the dataset.
\end{itemize}

Z-score is commonly used in detection of outliers (univariate analysis) in the dataset, by applying the arbitrarily selected thresholds. Selection of threshold has crucial effect on the reliability of the outlier detection, and it should be done by the domain expert. Also, such technique is sensitive to the data distribution; Z-scores assume that the data is normally distributed as they are using mean as a reference. Z-scores are sensitive to skewness (asymmetry) and kurtosis in the data. In the presence of significant skewness or heavy tails, z-scores may misclassify some data points as outliers or fail to identify true outliers. Finally, Z-scores assume that data points are independent of each other. If data points are correlated or exhibit serial dependence (e.g., time series data), z-scores may not be appropriate. Still, the negative effect of cross-correlation in time series data may be mitigated by using derivatives instead of the actual data-point values.

Multivariate outliers are data points that are outliers not in a single variable (uni-variate outlier) but rather in a multivariate context. Multivariate outliers are detect-ed by considering the relationships between two or more variables. A data point may not be considered an outlier when examining any individual variable but becomes an outlier when analyzing the combination of variables.

Multivariate outliers are important to identify because they can distort statistical analyses, affect the accuracy of predictive models, or carry unique information that may need further investigation. In the research presented in this paper, the following methods for multivariate (in this case, bivariate) outlier detection are: Mahalanobis distance, Principal Component Analysis (PCA), Isolation Forest and Hotelling's T-squared test. They are described in the remainder of this section.

\subsubsection{Mahalanobis distance}

Mahalanobis Distance \cite{Mahalanobis2018} is a method used for multivariate outlier detection. It is a measure of the distance between a data point and the center of a dataset in a multidimensional space, considering the variability and correlations of the data.
\begin{equation}
D(X) = \sqrt{(X-\mu)^{T}\Sigma^{-1}(X-\mu)}
\label{eq:mahalanobis}
\end{equation}
where:
\begin{itemize}
    \item $X$ represents the multivariate data point.
    \item $\mu$ is the mean vector.
    \item $\Sigma^{-1}$ is the inverse of the covariance matrix.
\end{itemize}

Mahalanobis Distance is robust to data with different scales and units. It can capture correlations and dependencies between variables, and it is applicable to high-dimensional datasets. One of the limitations of the method is that it assumes that the data follow a multivariate normal distribution, which may not always be the case.

\subsubsection{Principal Component Analysis (PCA)}

Principal Component Analysis (PCA) \cite{Wold1987} is a dimensionality reduction technique used in multivariate outlier detection. While PCA is primarily known for dimensionality reduction and data compression, it can also be applied to identify outliers in multivariate datasets.

PCA's primary goal is to transform a dataset with multiple correlated variables (features) into a new set of uncorrelated variables called principal components. These components are linear combinations of the original features and capture most of the variance in the data. In PCA, the first few principal components capture the majority of the variance in the dataset, while subsequent components capture less and less. Therefore, the first few components contain the most valuable information about the data. By reducing the dimensionality of the data with PCA, outliers can become more apparent in the reduced-dimensional space. PCA assumes that the data follow a Gaussian distribution, which may not always hold true for all datasets.

\subsubsection{Isolation Forest}

The Isolation Forest (IF) \cite{Liu2012} is an ensemble ML algorithm used for outlier detection. It operates by randomly selecting a feature and a random split point for that feature. It does this recursively until it isolates an outlier or reaches a specified depth in the tree.

The contamination parameter is used to control the proportion of expected outliers in the dataset. It represents the expected percentage of anomalies in the dataset. The contamination parameter controls the trade-off between precision and recall in outlier detection. A lower value results in stricter detection of outliers.

Some crucial advantages of IF are that it scales well to large datasets, it is capable of identifying anomalies in high-dimensional data, with relatively small tuning effort. Finally, it doesn't require assumptions about the data distribution.

However, there are some limitations. Isolation Forest assumes that anomalies are sparse and can be isolated with fewer splits. While this is often true for many datasets, it may not be the case for all. In some datasets, anomalies may not be isolated efficiently by random partitioning. The method is also sensitive to sampling as it relies on random sampling, which can lead to variation in results. Different random samples can produce slightly different isolation trees and, consequently, different outlier scores. Furthermore, IF treats features independently and doesn't consider the relationships or correlations between features. In some cases, outliers may only be detected by considering multivariate relationships. Thus, it is primarily designed for detecting individual outliers rather than identifying clusters of anomalies. Further-more, the method is sensitive to data scaling. It is important to pre-process the data appropriately, especially when features have different scales. The explainability is a challenge: Isolation Forest doesn't provide explanations or reasons for why a particular data point is considered an outlier. It only assigns anomaly scores based on path lengths in the trees.

\subsubsection{Hotelling's T-squared test}

Hotelling's T-squared test \cite{Hotelling1931} is a statistical method used for multivariate outlier detection and analysis.

Hotelling's T-squared test is designed to identify whether there are significant differences between the means of multivariate datasets. It provides a statistical measure of how far an observation is from the mean of the dataset in a multivariate space. The test compares the Mahalanobis distance (a measure of distance that accounts for correlations between variables) of each observation from the multivariate mean to a critical value derived from the F-distribution. Hotelling's T-squared test assumes that the data follows a multivariate normal distribution.

The alpha parameter in Hotelling's T-squared test refers to the significance level of the test. A lower alpha value means you are being more conservative and requiring stronger evidence to reject the null hypothesis.

\section{Implementation and discussion}

Bivariate detection of outliers is carried out on transmitted energy and outside temperature, where the latter is shifted one timestep in the future, because the decision on increasing the heat energy transmission in the next hour is made based on the temperature in the current hour. This is considered a simplification, because it assumes system inertia of 1 hour which is not actually correct. This simplification is made because only hourly series data is available for the period of analysis. Still, this simplification is reasonable because the effect of the decision is still embodied in the transmitted energy.

Both features are normally distributed which is very positive for great most of the methods for outlier detection which assume the Gaussian distribution. One exception is the concentration of zero data in transmitted energy, where in many data-points, no energy is being transmitted. This data is omitted from the analysis of outliers.
All the mentioned methods need to have some threshold defined, which, in the majority of the used methods, comes from a significant level value (often denoted as $\alpha$). Defining these values is somewhat of a subjective decision, and there is no universal way or guidelines to select these. There are recommendations for choosing the threshold and significant level, but they are also very dependent on the dataset for which the threshold is being defined, and expert knowledge of the domain from which the dataset was from. In the initial analysis, the authors used such recommendations. For the significant level, most often the recommendation is to use $\alpha$ = 0.05, which corresponds to a confidence level of 95\%.

In other cases, (e.g. Mahalanobis Distances) the recommendation is to use a threshold equal to 2 times the standard deviation of the data. However, when such thresholds were used, the outliers detect-ed by the methods were not very promising. According to the expert who analyzed the outliers detected by the methods, the number of outliers was much higher than what it should have been, and a significant portion of the identified outliers lacked a clear justification; the data associated with these outliers did not exhibit unusual characteristics. This indicated that most of the outliers identified by the methods were not true outliers but rather false positives. It implied that the chosen confidence level of 95\% ($\alpha$ = 0.05) was too low, and similarly, the threshold value calculated using the standard deviation was set too high.

Due to this observation, in the subsequent iteration, thresholds were approximated with experts examining the outcomes of outlier detection. The criteria for success were determined based on the expert's satisfaction with both the quantity and interpretability of the identified outliers. Once the expert deemed the results satisfactory, the thresholds were considered successful. Selected parameters were as follows: TMD=1 (Threshold parameter for MD method), TPCA=2 (Threshold parameter for PCA method), CIF=0.015 (Contamination parameter for IF method), AH=0.0007 (Alpha parameter for IF method).

However, despite the careful selection of thresholds, two methods yielded unsatisfactory results. The Univariate Outlier Detection method (Z-score) performed poorly, considering its focus on a single variable. However, it's worth noting that this method was employed merely as a benchmark, and its suboptimal performance was expected. The other method presenting challenges was Mahalanobis Distances. Regardless of the threshold settings, the identified outliers did not meet the expected interpretability. This issue warrants further investigation in future work. 
From results presented in Table 1, obtained from the three methods that were deemed successful, it is possible to draw certain conclusions.

\begin{table}[H]
\centering
\caption{Average number of outliers detected by PCA, Isolation Forest, and Hotelling T-test at each hour of the day, for both depicted seasons.}
\label{tab:hourly_outliers}
\begin{tabular}{lcc}
\toprule
\textbf{Hour} & \textbf{Average number of outliers} & \textbf{Average number of outliers} \\
              & \textbf{in Season 1}                & \textbf{in Season 2} \\
\midrule
07:00 & 3.6 & 6.6 \\
08:00 & 3.2 & 5.2 \\
09:00 & 3.6 & 4.0 \\
10:00 & 3.2 & 2.2 \\
11:00 & 3.6 & 3.0 \\
12:00 & 4.2 & 1.6 \\
13:00 & 3.8 & 1.4 \\
14:00 & 2.0 & 2.0 \\
22:00 & 6.8 & 5.2 \\
23:00 & 2.3 & 4.4 \\
\bottomrule
\end{tabular}
\end{table}

During season 1, there was a relatively consistent average number of outliers detect-ed by the methods for each hour, with the peak occurring at 22h. In contrast, season 2 exhibited varying numbers of detected outliers for each hour throughout the day. On average, the highest number of outliers appeared during the early morning hours (7h) and late-night hours (22h and 23h).

The increased number of outliers at 7h in the morning can be attributed to variations in the start time of the heating cycle by the operator at the heating plant. Some-times the cycle begins a bit later, and occasionally, the operator decides it's unnecessary to initiate heating at that specific time. This variability is reflected in the higher number of outliers detected at 7h. Similarly, during late-night hours (22h or 23h), the need for heating fluctuates based on the outside temperature and the operator's estimations.

\begin{figure}[H]
	\centering
	\includegraphics[width=0.7\textwidth]{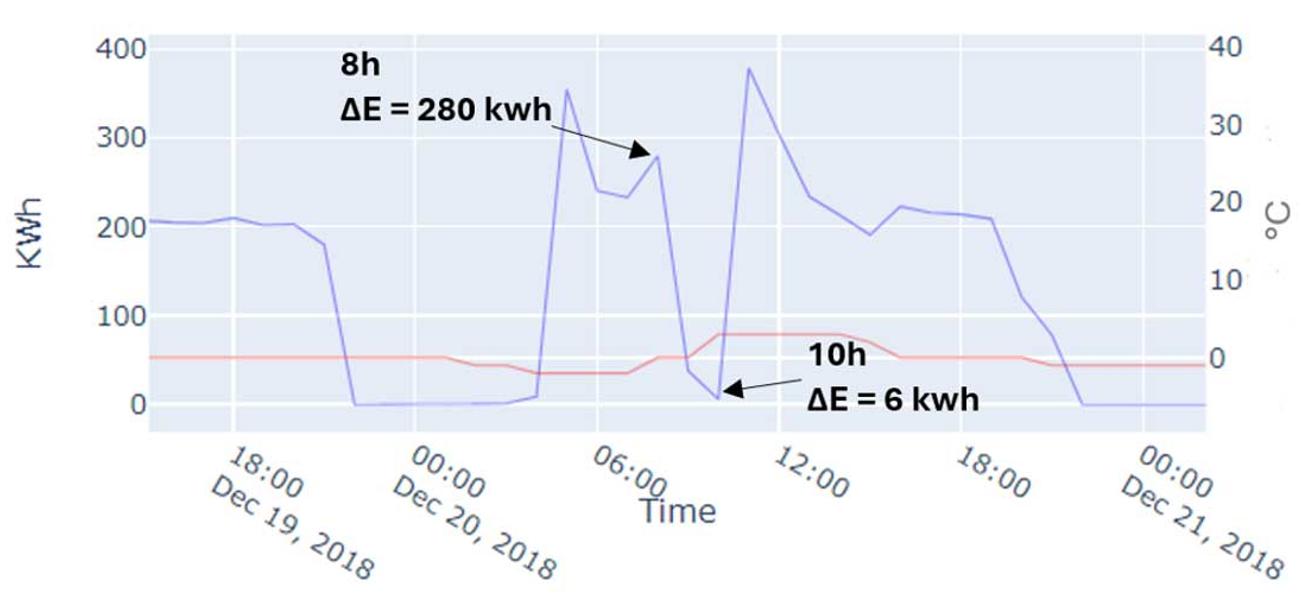}
	\caption{A sudden decrease of sent energy to the secondary supply line, due to exceeding the targeted temperature of the supply line in the previous hours.}
	\label{fig:fig2}
\end{figure}

There are instances where heating is deemed necessary and times when it is not, leading to a higher number of outliers during these hours. In addition, at times in the morning (9-11h) there appeared to be either no output energy or a minimal amount, even when the outside temperatures are low, and the heating system should be active (Figure 2). However, there is a straightforward explanation: during the early morning when the heating starts, the objective is to achieve the required temperature in the secondary supply line as fast as possible, to ensure customers receive adequate heating. Occasionally, too much energy is sent to the secondary supply line, causing the water temperature in that line to exceed the required level. Therefore, there is a significant reduction in the sent energy in the next hour(s), in order to lower the water temperature at the secondary line and thereby reach the required temperature level. While it was anticipated that the outlier detection methods might incorrectly label these segments as false positive outliers, the methods performed remarkably well by correctly recognizing that these instances are not outliers.

Overall, the methods displayed similar results, as all three noticed the majority of the outliers at the beginning of the heating season and at the end (Figure 3). Because these months of the winter are often warmer and the variation in outside temperature is more pronounced, the heating is also more unpredictable, which results in more outliers detected.

\begin{figure}[H]
	\centering
	\includegraphics[width=0.9\textwidth]{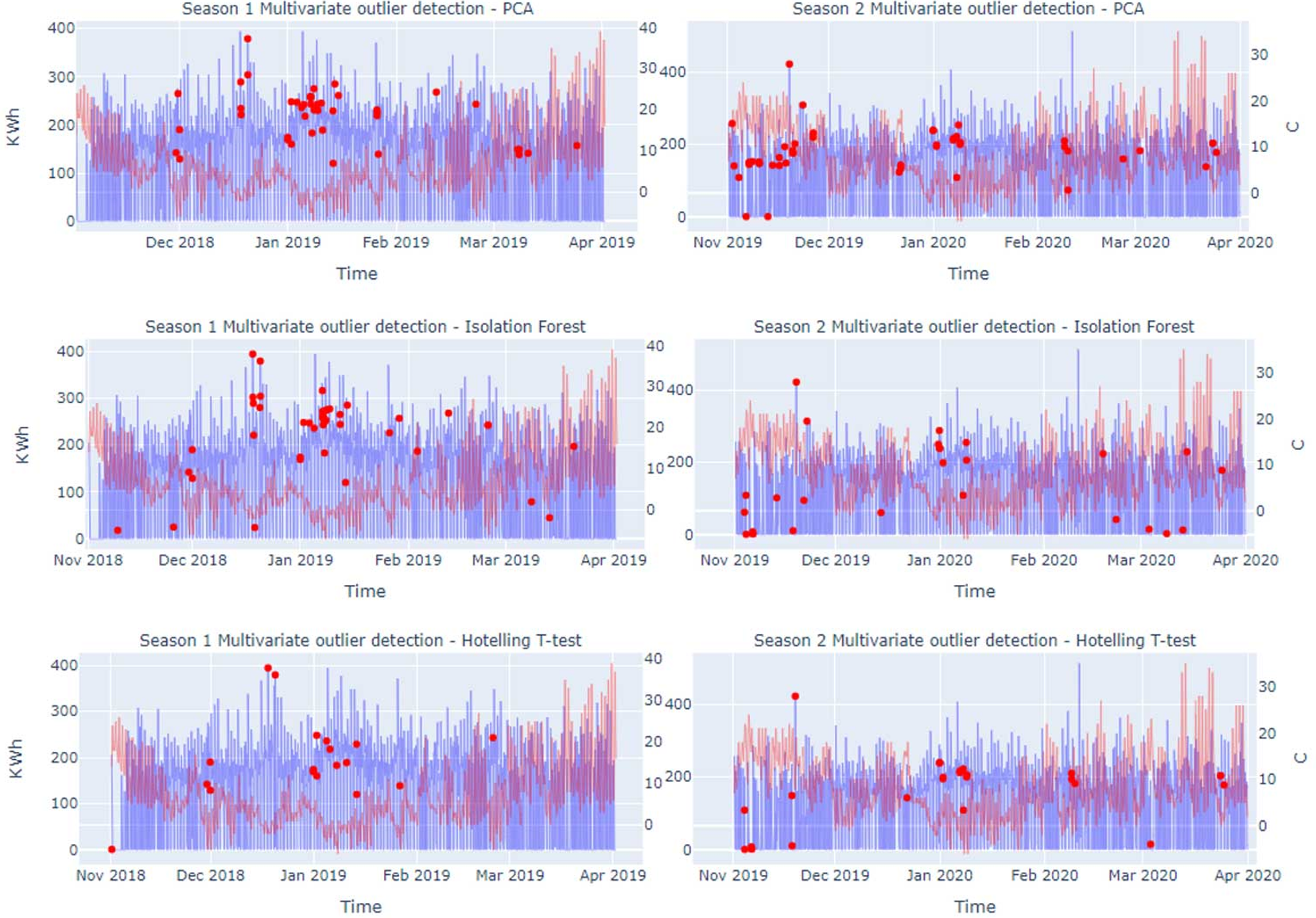}
	\caption{Detected outliers for each of the methods in both heating seasons.}
	\label{fig:fig3}
\end{figure}

However, most of the outliers are noticed at the beginning of January, which is odd at first, but understandable when looked more closely. Figure 4 shows that the majority of the outliers in the month of January are between the first and the 13th of January. Because this period is a holiday period of the year, the heating is increased as it is presumed that more users will be staying at home during the days, which is why more outliers are detected.

\begin{figure}[H]
	\centering
	\includegraphics[width=0.9\textwidth]{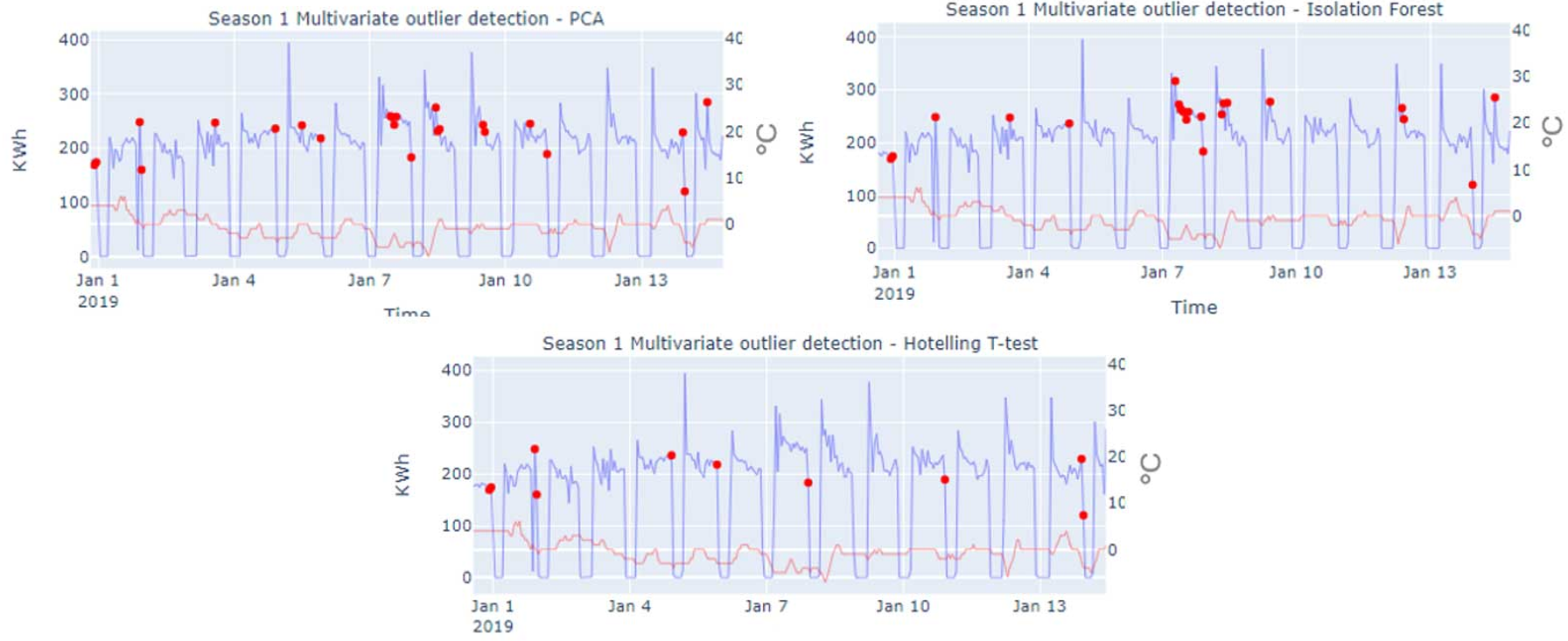}
	\caption{A closer look at the detected outliers in the first season in the holiday period of the year (01st Jan-13th Jan).}
	\label{fig:fig4}
\end{figure}

Even though the methods detected similar outliers, they did not always detect the exact same one, the methods worked relatively well, as commented by the expert. By closely analyzing the results and discussing them with the domain expert, it was concluded that the best choice is to use two methods (PCA, Isolation Forest) in con-junction, since Hotelling T-test results were more conservative. The outliers which are detected by both of these methods would be deemed the right outliers, all the others would be false positives.

\section{Conclusion}

In this preliminary research, we have identified the strengths and weaknesses of different methods for the bivariate detection of outliers in DHS operation, based on the features of transmitted heat energy and ambient temperature. PCA and Isolation Forest methods have been adopted as reference ones, producing satisfactory results. The research conclusions are still considered as weak since the adoption of the methods is based on the inspection of the detected outliers by the expert.

In the future, the research will take the direction of considering more features in a multivariate analysis, namely other relevant weather parameters, such as solar irradiance, wind strength and direction. One of the next steps will be to test the effectiveness of the different boosting methods within the multivariate outlier detection problem. For example, PCA can be used to reduce the dimensionality to transform the multivariate to bivariate problem, where other methods can be used to solve it, such as Isolation Forest. The performance measurement of the conventional methods still remains the problem as it requires quite a significant effort by the experts who validate the results manually. Efficiency of the validation process, at least in the aspect of method comparison can be somewhat improved by implementing the dynamic selection of the values of parameters/thresholds for different methods based on the defined fixed number of outliers to be detected for all methods.

Feasibility of the supervised anomaly detection approach will be investigated, based on expert annotated anomalies. Such an approach would enable explicit and direct performance measurement, and it will facilitate the explainability. Also, other methods, namely Deep Learning based ones will be tested. Finally, the research will deliver the software application for walk-forward detection of outliers which are expected to pinpoint ineffective or inefficient heating as well as the fault detection.

\bibliographystyle{unsrt}
\bibliography{references}

\end{document}